\documentclass[journal]{IEEEtran}
\usepackage{lipsum} % Just for the demo, you can safely remove it. 
\usepackage[tight,footnotesize]{subfigure}
\usepackage{enumitem} 
\usepackage{float}
\usepackage{multicol,multienum}
\usepackage{breqn}
\usepackage{stfloats}
\usepackage{multirow}
\usepackage{diagbox}
\usepackage{hyperref}
\usepackage{bm}
\usepackage{cite}
\usepackage{graphicx}
\usepackage{tabularx}
\usepackage[ruled,vlined]{algorithm2e} 
\usepackage{booktabs}

\usepackage[tight,footnotesize]{subfigure}
\usepackage{url}
\usepackage{doi}
\usepackage{amssymb}
\usepackage{xcolor}
\ifCLASSINFOpdf

\else

\fi

\begin{document}

\title{Deep Anomaly Detection for Time-series Data in Industrial IoT: A Communication-Efficient On-device Federated Learning Approach}

%\title{ Communication-Efficient Federated Learning for Anomaly Detection in  Industrial Internet of Things} 

\author{Yi~Liu,
 Sahil Garg,~\IEEEmembership{Member,~IEEE,} Jiangtian Nie, Yang Zhang,  Zehui Xiong,  Jiawen Kang, \\M. Shamim Hossain,~\IEEEmembership{Senior Member,~IEEE}
 
%\thanks{Y. Liu is  with the School of Data Science and Technology, Heilongjiang University, Harbin, China (email: 97liuyi@ieee.org). S. Garg is with the Electrical Engineering Department, \'Ecole de technologie sup\'erieure, Universit\'e du Qu\'ebec, Montr\'eal, QC H3C 1K3, Canada. (e-mail: sahil.garg@ieee.org). J. Nie, Z. Xiong, and J. Kang are with the School of Computer Science and Engineering, Nanyang Technological University, Singapore (emails: jnie001@e.ntu.edu.sg, ZXIONG002@e.ntu.edu.sg, kavinkang@ntu.edu.sg). Y. Zhang is Hubei Key Laboratory of Transportation Internet of Things, School of Computer Science and Technology, Wuhan University of Technology, China  (email:yangzhang@whut.edu.cn).  M. Shamim Hossain is with the Department of Software Engineering, College of Computer and Information Sciences, King Saud University, Riyadh 11543, Saudi Arabia (E-mail: mshossain@ksu.edu.sa)}}

\thanks{
This research is supported by   Alibaba Group through Alibaba Innovative Research (AIR) Program and Alibaba-NTU Singapore Joint Research Institute (JRI), NTU, Singapore, and the Open Research Project of the State Key Laboratory of Industrial Control Technology, Zhejiang University, China (No.ICT20044), National Natural Science Foundation of China (Grant No. 51806157), Young Innovation Talents Project in Higher Education of Guangdong Province, China (Grant No. 2018KQNCX333). The authors extend their appreciation to the Researchers Supporting Project number (RSP-2020/32), King Saud University, Riyadh, Saudi Arabia for funding this work. (Corresponding author: Jiawen Kang)

Y. Liu is  with the School of Data Science and Technology, Heilongjiang University, Harbin, China (email: 97liuyi@ieee.org). S. Garg is with the Electrical Engineering Department, \'Ecole de technologie sup\'erieure, Universit\'e du Qu\'ebec, Montr\'eal, Canada. (e-mail: sahil.garg@ieee.org). J. Nie is with Energy Research Institute, Interdisciplinary Graduate Programme and School of Computer Science and Engineering, Nanyang Technological University, Singapore. (e-mail: jnie001@e.ntu.edu.sg). Y. Zhang is Hubei Key Laboratory of Transportation Internet of Things, School of Computer Science and Technology, Wuhan University of Technology, China  (email:yangzhang@whut.edu.cn).  Z. Xiong is with  Alibaba-NTU Joint Research Institute and also School of Computer Science and Engineering, NTU, Singapore. (Email: zxiong002@e.ntu.edu.sg). J. Kang is with Energy Research Institute, Nanyang Technological University, Singapore (e-mail: kavinkang@ntu.edu.sg). M. Shamim Hossain is with the Department of Software Engineering, College of Computer and Information Sciences, King Saud University,  Saudi Arabia (E-mail: mshossain@ksu.edu.sa).}

%Y. Liu is  with the School of Data Science and Technology, Heilongjiang University, Harbin, China (email: 97liuyi@ieee.org). S. Garg is with the Electrical Engineering Department, \'Ecole de technologie sup\'erieure, Universit\'e du Qu\'ebec, Montr\'eal, QC H3C 1K3, Canada. (e-mail: sahil.garg@ieee.org). J. Nie is with Energy Research Institute, Interdisciplinary Graduate Programme and School of Computer Science and Engineering, Nanyang Technological University, Singapore. (e-mail: jnie001@e.ntu.edu.sg). Y. Zhang is Hubei Key Laboratory of Transportation Internet of Things, School of Computer Science and Technology, Wuhan University of Technology, China  (email:yangzhang@whut.edu.cn).  Z. Xiong is with  Alibaba-NTU Joint Research Institute and also School of Computer Science and Engineering, NTU, Singapore. (Email: zxiong002@e.ntu.edu.sg). J. Kang is with Energy Research Institute, Nanyang Technological University, Singapore (e-mail: kavinkang@ntu.edu.sg). M. Shamim Hossain is with the Department of Software Engineering, College of Computer and Information Sciences, King Saud University, Riyadh 11543, Saudi Arabia (E-mail: mshossain@ksu.edu.sa)}
}

\markboth{IEEE Internet of Things Journal}%
{Shell \MakeLowercase{\textit{et al.}}: Bare Demo of IEEEtran.cls for IEEE Journals}

\maketitle
\begin{abstract}
Since edge device failures (i.e., anomalies) seriously affect the production of industrial products in Industrial IoT (IIoT), accurately and timely detecting anomalies is becoming increasingly important. Furthermore, data collected by the edge device may contain the user's private data, which is challenging the current detection approaches as user privacy is calling for the public concern in recent years. With this focus, this paper proposes a new communication-efficient on-device federated learning (FL)-based deep anomaly detection framework for sensing time-series data in IIoT. Specifically, we first introduce a FL framework to enable decentralized edge devices to collaboratively train an anomaly detection model, which can improve its generalization ability.  {Second, we propose an Attention Mechanism-based Convolutional Neural Network-Long Short Term Memory (AMCNN-LSTM) model to accurately detect anomalies.} The AMCNN-LSTM model uses attention mechanism-based CNN units to capture important fine-grained features, thereby preventing memory loss and gradient dispersion problems. Furthermore, this model retains the advantages of LSTM unit in predicting time series data. Third, to adapt the proposed framework to the timeliness of industrial anomaly detection, we propose a gradient compression mechanism based on Top-\textit{k} selection to improve communication efficiency.  {Extensive experiment studies on four real-world datasets demonstrate that the proposed framework can accurately and timely detect anomalies and also reduce the communication overhead by 50\% compared to the federated learning framework that does not use gradient compression scheme.}

%Experimental results show that this mechanism can compress the gradient by 300 times without losing accuracy. Extensive experiment studies on four real-world datasets demonstrate that the proposed framework can accurately and timely detect anomalies and also reduce about 50\% communication overhead of traditional frameworks.
\end{abstract}


\IEEEpeerreviewmaketitle

\section{Introduction}
\IEEEPARstart{T}{he} widespread deployment of edge devices in the Industrial Internet of Things (IIoT) paradigm has spawned a variety of emerging applications with edge computing, such as smart manufacturing, intelligent transportation, and intelligent logistics \cite{peng2020deep}.  {The edge devices provide powerful computation resources to enable real-time, flexible, and quick decision making for the IIoT applications, which has greatly promoted the development of Industry 4.0 \cite{ref-1}.} However, the IIoT applications are suffering from critical security risks caused by abnormal IIoT nodes which hinders the rapid development of IIoT. For example, in smart manufacturing scenarios, industrial devices acting as IIoT nodes, e.g., engines with sensors, that have abnormal behaviors (e.g., abnormal traffic and irregular reporting frequency) may cause industrial production interruption thus resulting in huge economic losses for factories \cite{ref-4,peng2016toward}. Edge devices (e.g., industrial robots), generally collect sensing data from IIoT nodes, especially time-series data, to analyze and capture the behaviors and operating condition of IIoT nodes by edge computing \cite{ref-3}. Therefore, these sensing time series data can be used to detect the anomaly behaviors of IIoT nodes \cite{ref-2}. 

To solve the abnormality problems from IIoT devices, typical methods are to perform abnormal detection for the affected IIoT devices\cite{8758843,ref-5, garg2020multi,garg2020abc}. Previous work focused on utilizing deep anomaly detection (DAD) \cite{ref-13} approaches to detect abnormal behaviors of IIoT devices by analyzing sensing time series data.  {DAD techniques can learn hierarchical discriminative features from historical time-series data. In \cite{ref-14,ref-46,ref-47}, the authors proposed a Long Short Term Memory (LSTM) networks-based deep learning model to achieve anomaly detection in sensing time series data.} %The authors in \cite{ref-15} proposed a DAD framework, called DeepAD, that leverages deep learning-based time-series forecasting models to detect anomalies more accurately. 
Munir \textit{et al.} in \cite{ref-16} proposed a novel DAD approach, called DeepAnT, to achieve anomaly detection by utilizing deep Convolutional Neural Network (CNN) to predict anomaly value. Although the existing DAD approaches have achieved success in anomaly detection, they cannot be directly applied to the IIoT scenarios with distributed edge devices for timely and accurate anomaly detection. The reasons are two-fold: (i) the most of detection models are not flexible enough in traditional approaches,  the edge devices lack  dynamic and automatically updated detection models for different scenarios,  and hence the models fail to accurately predict frequently updated time-series data \cite{ref-5}; (2) due to privacy concerns, the edge devices are not willing to share their own collected time-series data with each other, thus the data exists in the form of ``islands.'' The data islands significantly degrade the performance of anomaly detection.  {Furthermore, it is often overlooked that the data may contain sensitive private information, which leads to potential privacy leakage. There are some privacy issues in the anomaly detection context. For example, the anomaly detection model will reveal the patient's heart disease history when detecting the patient's abnormal pulse \cite{ref-44,ref-45}.}% This is the reason why these DAD models cannot be applied to distributed edge devices.

To address the above challenges, a promising on-device privacy-preserving distributed machine learning paradigm, called on-device federated learning (FL), was proposed for edge devices to train a global DAD model while keeps the training datasets locally without sharing raw training data \cite{ref-6}. Such a framework allows edge devices to collaboratively train an on-device DAD model without compromising privacy.  {For example, the authors in \cite{ref-17} proposed an on-device FL-based approach to achieve collaborative anomaly detection.} Tsukada \textit{et al.} in \cite{ref-18} utilized FL framework to propose a Backpropagation Neural Networks (i.e., BPNNs) based approaches for anomaly detection. However, previous researches ignore the communication overhead during model training using  {FL} among large-scale edge devices. Expensive communication overhead may cause excessive overhead and long convergence time for edge devices so that the on-device DAD model cannot quickly detect anomalies. Therefore, it is necessary to develop a communication-efficient on-device FL framework to achieve accurate and timely anomaly detection for edge devices.

In this paper, we propose a communication-efficient on-device FL framework that leverages an attention mechanism-based CNN-LSTM (AMCNN-LSTM) model to achieve accurate and timely anomaly detection for edge devices.  {First, we introduce a FL framework to enable distributed edge devices to collaboratively train a global DAD model without compromising privacy.}  {Second, we propose an AMCNN-LSTM model to detect anomalies.} Specifically, we use attention-based CNNs to extract fine-grained features of historical observation-sensing time-series data and use LSTM modules for time-series prediction. Such a model can prevent memory loss and gradient dispersion problems. Third, to further improve the communication efficiency of the proposed framework, we propose a gradient compression mechanism based on Top-\textit{k} selection to reduce the number of gradients uploaded by edge devices. We evaluate the proposed framework on four real-world datasets: power demand, space shuttle, ECG, and engine. Experimental results show that the proposed framework can achieve high-efficiency communication and achieve accurate and timely anomaly detection. The contributions of this paper are summarized as follows:
\begin{itemize}
	\item We introduce a federated learning framework to develop an on-device collaborative deep anomaly detection model for edge devices in IIoT.
	\item We propose an attention mechanism-based CNN-LSTM model to detect anomalies, which uses CNN to capture the fine-grained features of time series data and uses LSTM module to accurately and timely detect anomalies.
	\item We propose a Top-\textit{k} selection-based gradient compression scheme to improve the proposed framework's communication efficiency. Such a scheme decreases communication overhead by reducing the exchanged gradient parameters between the edge devices and the cloud aggregator.
	\item We conduct extensive experiments on four real-world datasets to demonstrate that the proposed framework can accurately detect anomalies with low communication overhead.
\end{itemize}
\section{Related Work}
\subsection{Deep Anomaly Detection}
Deep Anomaly Detection (DAD) has always been a hot issue in IIoT, which serves as a function of detecting anomalies. Previous researches about DAD generally can be divided into three categories: \textit{supervised DAD}, \textit{semi-supervised DAD}, and \textit{unsupervised DAD} approaches.

\textbf{Supervised Deep Anomaly Detection:}
Supervised deep anomaly detection typically uses the labels of normal and abnormal data to train a deep-supervised binary or multi-class classifier. For example, Erfani \textit{et al.} in \cite{ref-19} proposed a supervised Support Vector Machine (SVM) classifier for high-dimensional data to classify normal and abnormal data. Despite the success of supervised DAD methods in anomaly detection, these methods are not as popular as semi-supervised or unsupervised methods due to the lack of labeled training data \cite{ref-15}. Furthermore, the supervised DAD method has poor performance for data with class imbalance (the total number of positive class data is much larger than the total number of negative class data) \cite{ref-14}.

\textbf{Semi-supervised Deep Anomaly Detection:} 
Since normal instances are easier to obtain the labels than that of anomalies, semi-supervised DAD techniques are proposed to utilize a single (normally positive class) existing label to separate outliers \cite{ref-13}. For example, Wulsin \textit{et al.} in \cite{ref-20} applied Deep Belief Nets (DBNs) in a semi-supervised paradigm to model Electroencephalogram (EEG) waveforms for classification and anomaly detection. The semi-supervised DBN performance is comparable to the standard classifier on EEG dataset. %Akcay \textit{et al.} in \cite{ref-21} proposed a novel DAD framework that uses adversarial training to train DAD classifiers in a semi-supervised manner. 
The semi-supervised DAD approach is popular because it can use only a single class of labels to detect anomalies.

\textbf{Unsupervised Deep Anomaly Detection:}
Unsupervised deep anomaly detection techniques use the inherent properties of data instances to detect outliers \cite{ref-13}. For example, Zong \textit{et al.} in \cite{ref-22} proposed a deep automatic coding Gaussian Mixture Model (DAGMM) for unsupervised anomaly detection. Schlegl \textit{et al.} in \cite{ref-23} proposed a deep convolutional generative adversarial network, called AnoGAN, which detects abnormal anatomical images by learning a variety of normal anatomical images. They trained such a model in an unsupervised manner. Unsupervised DAD is widely used since it does not require the characteristics of labeled training data. 

%\subsection{On-device Federated Learning}
%Due to the rapid development of edge intelligence technology, more and more deep learning models are applied to edge devices, which has spawned a new edge intelligence paradigm called on-device federated learning. The on-device federated learning allows edge devices to train machine learning models on local datasets without sharing the raw data \cite{ref-24}. For example, the author in \cite{ref-26} applied a deep neural network to edge devices to build a large visual recognition model. The authors in \cite{ref-27} deployed deep learning techniques to devices to improve the accuracy of speech recognition. Furthermore, to enable complex on-device deep learning models, many communication optimization algorithms have been proposed. Jakub \textit{et al.} in \cite{ref-24} proposed a federated optimization algorithm for on-device learning to improve communication efficiency. The authors in \cite{ref-25} proposed an augmentation algorithm under non-iid private data to improve communication efficiency for on-device learning.
%As a result, in this paper, we propose an on-device attention mechanism-based CNN-LSTM model to detect anomalies on edge devices.

\subsection{Communication-Efficient Federated Learning}
Google proposed a privacy-preserving distributed machine learning framework, called FL, to train machine learning models without compromising privacy \cite{ref-28}. Inspired by this framework, different edge devices can contribute to the global model training while keeping the training data locally. However, communication overhead is the bottleneck of FL being widely used in IIoT \cite{ref-6}. Previous work has focused on designing efficient stochastic gradient descent (SGD) algorithms and using model compression to reduce the communication overhead of FL. Agarwal \textit{et al.} in \cite{ref-30} proposed an efficient cpSGD algorithm to achieve communication-efficient FL. Reisizadeh \textit{et al.} in \cite{ref-29} used Periodic Averaging and Quantization methods to design a communication-efficient FL framework. Jeong \textit{et al.} in \cite{ref-25} proposed a federated model distillation method to reduce the communication overhead of FL. 

However, the above methods do not substantially reduce the number of gradients exchanged between edge devices and the cloud aggregator. The fact is that a large number of gradients exchanged between the edge devices and the cloud aggregator may cause excessive communication overhead for FL \cite{ref-11}. Therefore, in this paper, we propose a Top-\textit{k} selection-based gradient compression scheme to improve the communication efficiency of FL.

\section{Preliminary}
In this section, we briefly  introduce anomalies, federated deep learning, and gradient compression as follows.
%\begin{figure}[!t]
%	\centering
%	\includegraphics[width=0.5\linewidth]{figs/fig-2.png}
%	\caption{Illustration of anomalies in two-dimensional dataset.}
%	\label{fig-1}
%\end{figure}
\subsection{Anomalies}
 {In statistics, anomalies (also called outliers and abnormalities) are the data points that are significantly different from other observations \cite{ref-13}. We assume that $N_1$, $N_2$, and $N_3$ are regions composed of most observations, so they are regarded as normal data instance regions. If data points $O_1$ and $O_2$ are far from these regions, they can be classified as anomalies. To define anomalies more formally, we assume that an \textit{n}-dimensional dataset $\vec{x}_{i}=\left(x_{i, 1}, \ldots, x_{i, n}\right)$ follows a normal distribution and its mean ${\mu _{\rm{j}}}$ and variance ${\sigma _j}$ for each dimension where $i \in\{1, \ldots, m\}$ and $j \in\{1, \ldots, n\}$. Specifically, for $j \in\{1, \ldots, n\}$, under the assumption of the normal distribution, we have:}
%	 $\mu _{j}=\sum_{i=1}^{m} x_{i, j} / m,\;\sigma_{j}^{2}=\sum_{i=1}^{m}\left(x_{i, j}-\mu_{j}\right)^{2} / m$,}
 {
\begin{equation}
\mu _{j}=\sum_{i=1}^{m} x_{i, j} / m,\;\sigma_{j}^{2}=\sum_{i=1}^{m}\left(x_{i, j}-\mu_{j}\right)^{2} / m,
\end{equation}}
if there is a new vector $\vec{x}$, the probability $p (\vec{x})$ of anomaly can be calculated as follows:
\begin{equation}
p(\vec{x})=\prod_{j=1}^{n} p\left(x_{j} ; \mu_{j}, \sigma_{j}^{2}\right)=\prod_{j=1}^{n} \frac{1}{\sqrt{2 \pi} \sigma_{j}} \exp \left(-\frac{\left(x_{j}-\mu_{j}\right)^{2}}{2 \sigma_{j}^{2}}\right).
\end{equation}

%We can then judge whether vector $\vec{x}$ belongs to an anomaly according to the probability value for industrial big data \cite{ref-13}.

%In statistics, anomalies (also called outliers and abnormalities) are data points that are significantly different from other observations \cite{ref-13}. As shown in Fig. \ref{fig-1}, $N_1$, $N_2$, and $N_3$ are regions composed of most observations, so they are regarded as normal data instance regions. Therefore, data points $O_1$ and $O_2$ are far from these regions can be regarded as anomalies. To define anomalies more formally, we assume that an \textit{n}-dimensional dataset $\vec{x}_{i}=\left(x_{i, 1}, \cdots, x_{i, n}\right)$ follows a normal distribution and its mean ${\mu _{\rm{j}}}$ and variance ${\sigma _j}$ for each dimension where $i \in\{1, \cdots, m\}$ and $j \in\{1, \cdots, n\}$. Specifically, for $j \in\{1, \cdots, n\}$, under the assumption of normal distribution, we have:
%\begin{equation}
%\mu _{j}=\sum_{i=1}^{m} x_{i, j} / m,\;\sigma_{j}^{2}=\sum_{i=1}^{m}\left(x_{i, j}-\mu_{j}\right)^{2} / m.
%\end{equation}
%If there is a new data $\vec{x}$, the probability $p (\vec{x})$ of anomaly can be calculated as follows:
%\begin{equation}
%p(\vec{x})=\prod_{j=1}^{n} p\left(x_{j} ; \mu_{j}, \sigma_{j}^{2}\right)=\prod_{j=1}^{n} \frac{1}{\sqrt{2 \pi} \sigma_{j}} \exp \left(-\frac{\left(x_{j}-\mu_{j}\right)^{2}}{2 \sigma_{j}^{2}}\right).
%\end{equation}
We can then judge whether vector $\vec{x}$ belongs to an anomaly according to the probability value.
\subsection{Federated Learning}
Traditional distributed deep learning techniques require a certain amount of private data to be aggregated and analyzed at central servers (e.g., cloud servers) during the model training phase by using distributed stochastic gradient descent {(D-SGD)} algorithm \cite{ref-12}. Such the training process suffers from potential data privacy leakage risks for IIoT devices. To address such privacy challenges, a collaboratively distributed deep learning paradigm, called federated deep learning, was proposed for edge devices to train a global model while keeping the training datasets locally without sharing raw training data \cite{ref-6}. The procedure of FL is divided into three phases: the initialization phase, the aggregation phase, and the update phase. In the initialization phase, we consider that FL with $N$ edge devices and a parameter aggregator, i.e.,  a cloud aggregator,  distributes a pre-trained global model ${\omega _{t}}$ on the public datasets (e.g., MNIST \cite{ref-31}, CIFAR-10 \cite{ref-32}) to each edge devices. Following that, each device uses local dataset ${\mathcal{D}_k}$ of size $D_k$ to train and improve the current global model ${\omega _t}$ in each iteration. In the aggregation phase, the cloud aggregator collects local gradients uploaded by the edge nodes (i.e., edge devices). To do so, the local loss function to be optimized is defined as follows:  
\begin{equation}
\mathop {\min }\limits_{x \in {\mathbb{R}^d}} {F_k}(x) = \frac{1}{{{D_k}}}\sum\nolimits_{i \in {D_k}} { {\mathbb{E}_{{z_i} \sim {D_k}}}} f(x;{z_i}) + \lambda h(x),
\end{equation}
where $f( \cdot ; \cdot )$ is the local loss function for edge device $k$, $\forall \lambda  \in [0,1]$, $h( \cdot )$ is a regularizer function for edge device $k$, and $\forall i \in [1,\cdots,n],{z_i}$ is sampled from the local dataset $\mathcal{D}_k$ on the $k$ device. In the update phase, the cloud aggregator uses Federated Averaging (FedAVG) algorithm \cite{ref-28} to obtain a new global model $\omega_{t + 1}$ for the next iteration, thus we have:
\begin{equation}
{\omega _{t + 1}} \leftarrow {\omega _t} + \frac{1}{n}\sum\limits_{n = 1}^N {F_{t + 1}^n},
\end{equation}
where $\sum\limits_{n = 1}^N {F_{t + 1}^n} $ denotes model updates aggregation and $\frac{1}{n}\sum\limits_{n = 1}^N {F_{t + 1}^n} $ denotes the average aggregation (i.e., FedAVG algorithm). Both the edge devices and the cloud aggregator repeat the above process until the global model reaches convergence. This paradigm significantly reduces the risks of privacy leakage by decoupling the model training from direct access to the raw training data on edge nodes.

\subsection{Gradient Compression}
Large-scale FL training requires significant communication bandwidth for gradient exchange, which limits the scalability of multi-nodes training \cite{ref-9}. In this context, Lin \textit{et al.} in \cite{ref-9} stated that 99.9\% of the gradient exchange in D-SGD is redundant. To avoid expensive communication bandwidth limiting large-scale distributed training, gradient compression is proposed to greatly reduce communication bandwidth. Researchers generally use gradient quantization \cite{ref-33} and gradient sparsification \cite{ref-34} to achieve gradient compression. Gradient quantization reduces communication bandwidth by quantizing gradients to low-precision values. Gradient sparsification uses threshold quantization to reduce communication bandwidth. 

For a fully connected (FC) layer in a deep neural network, we have: $b=f(W*a+v)$, where $a$ is the input, $v$ is the bias, $W$ is the weight, $f$ is the nonlinear mapping, and $b$ is the output. This formula is the most basic operation in a neural network. For each specific neuron $i$, the above formula can be simplified to the following: $b_{i}=\mathrm{ReLU}\left(\sum_{j=0}^{n-1} W_{i j} a_{j}\right)$, Where $\mathrm{ReLU}$ is the activation function. Gradient compression compresses the corresponding weight matrix into a sparse matrix, and hence the corresponding formula is given as follows:
\begin{equation}
b_{i}=\mathrm{ReLU}\left(\sum_{j \in X_{i} \cap Y} \mathrm{Sparse}\left[I_{i j}\right] a_{j}\right),
\end{equation}
where $\sum_{j \in X_{i} \cap Y} S\left[I_{i j}\right]$ represents the compressed weight matrix and $i, j$ represent the position information of the gradient in the weight matrix $W$. Such a method reduces the communication overhead through sparsing the gradient in the weight matrix $W$.
\begin{figure}[!t]
	\centering
	\includegraphics[width=1\linewidth]{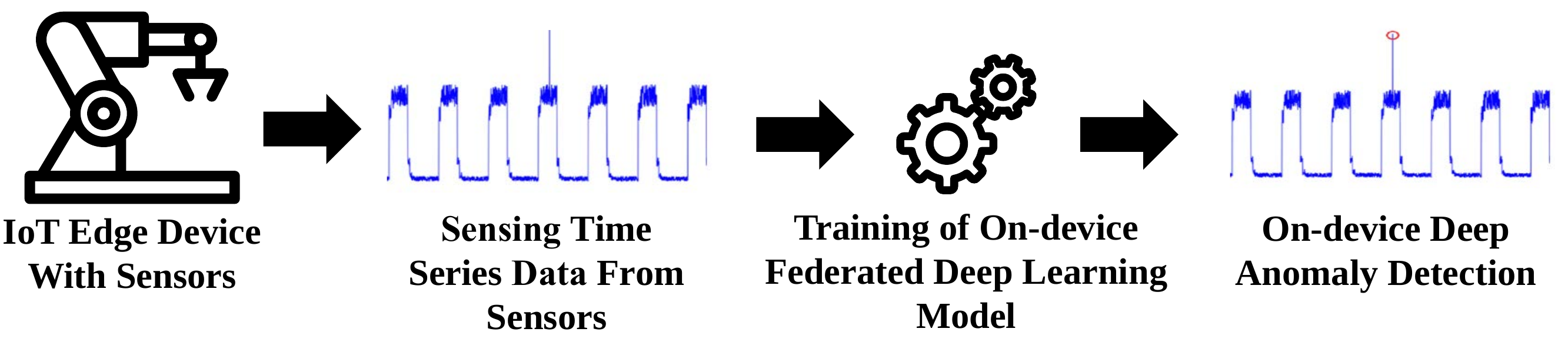}
	\caption{The workflow of the on-device deep anomaly detection in IIoT.}
	\label{fig-2}
	\vspace*{-3mm}
\end{figure}
\section{System Model}
We consider the generic setting for on-device deep anomaly detection in IIoT, where a cloud aggregator and edge devices work collaboratively to train a DAD model by using a given training algorithm (e.g., LSTM) for a specific task (i.e., anomaly detection task), as illustrated in Fig. \ref{fig-2}. The edge devices train a shared global model locally on their own local dataset (i.e., sensing time series data from IIoT nodes) and upload their model updates (i.e., gradients) to the cloud aggregator. The cloud aggregator uses the FedAVG algorithm or other aggregation algorithms to aggregate these model updates and obtains a new global model. In the end, the edge devices will receive the new global model sent by the cloud aggregator and use it to achieve accurate and timely anomaly detection.
\subsection{System Model Limitations}
 {The proposed framework focus on a DAD model learning task involving $N$ distributed edge devices and a cloud aggregator.} In this context, this framework has two limitations:: \textit{missing labels} and \textit{communication overhead}.

For missing-label limitation, we assume that the labels of the training sample with proportion $p\;(0<p<1)$ are missing. The lack of the label of the sample will cause the problem of class imbalance, thereby reducing the accuracy of DAD model. For communication-overhead limitation, we consider that there exists an excessive communication overhead when a large number of gradients exchanged between the edge devices and the cloud aggregator, which may make the model fail to converge \cite{ref-25}.

The above restrictions hinder the deployment of DAD model in edge devices, which motivates us to develop a communication-efficient FL-based unsupervised DAD framework to achieve accurate and timely anomaly detection.
\begin{figure}[!t]
	\centering
	\includegraphics[width=1\linewidth]{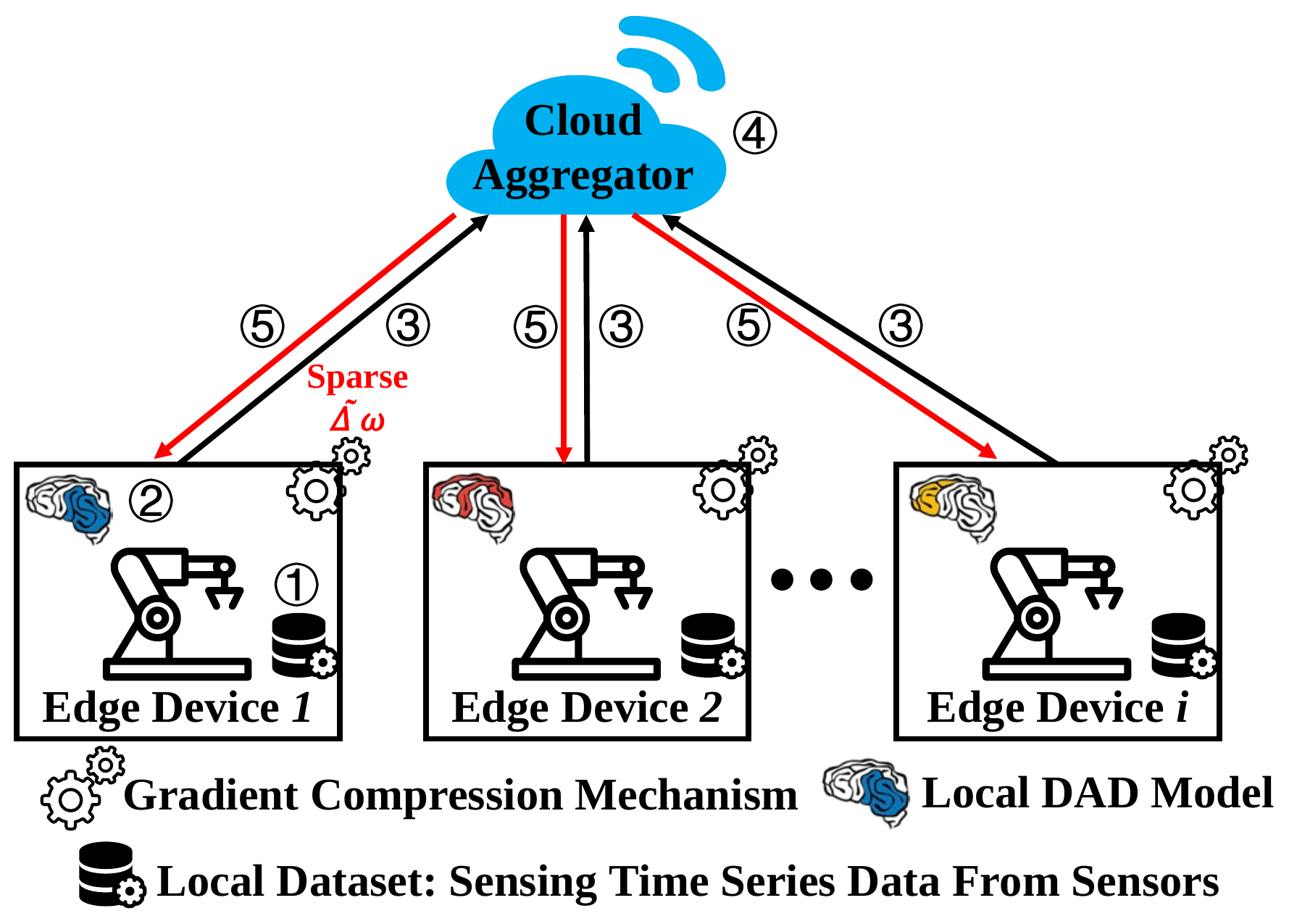}
	\caption{The overview of on-device communication-efficient deep anomaly detection framework in IIoT. This framework's workflow consists of five steps, as follows: (i) The edge device uses the sensing time series data collected from IIoT nodes as a local dataset (as shown in \textcircled{1}). (ii) The edge device performs local model (i.e., AMCNN-LSTM model) training on the local dataset (as shown in \textcircled{2}). (iii) The edge device uploads the sparse gradients $\tilde \Delta \omega $ to the cloud aggregator by using a gradient compression mechanism (as shown in \textcircled{3}). (iv) The cloud aggregator obtains a new global model by aggregating the sparse gradients uploaded by the edge device (as shown in \textcircled{4}). (v) The cloud aggregator sends the new global model to each edge device. The above steps are executed cyclically until the global model reaches optimal convergence (as shown in \textcircled{5}). Decentralized devices can use this optimal global model to perform anomaly detection tasks.}
	\label{fig-3}
	\vspace*{-5mm}
\end{figure}
\subsection{The Proposed Framework}
We consider an on-device communication-efficient deep anomaly detection framework that involves multiple edge devices for collaborative model training in IIoT, as illustrated in Fig. \ref{fig-3}. In particular, this framework consists of a cloud aggregator and edge devices. Furthermore, the proposed framework also includes two mechanisms: an anomaly detection mechanism and a gradient compression mechanism. More details are described as follows:
\begin{itemize}
	\item \textbf{\textit{Cloud Aggregator}:} The cloud aggregator is generally a cloud server with strong computing power and rich computing resources. The cloud aggregator contains two functions: (1) initializes the global model and sends the global model to the all edge devices; (2) aggregates the gradients uploaded by the edge devices until the model converges.
	\item \textbf{\textit{Edge Devices}:} Edge devices are generally agents and clients, such as whirlpool, wind turbine, and vehicle, which contain local models and functional mechanisms (see below for more details).  Each edge device uses the local dataset (i.e., sensing time series data from IIoT nodes) to train the global model sent by the cloud aggregator and uploads the gradients to the cloud aggregator until the global model converges. The local model is deployed in the edge device, and it can perform anomaly detection. In this paper, we use AMCNN-LSTM model to detect anomalies, which uses CNN to capture the fine-grained features of sensing time series data and uses LSTM module to accurately and timely detect anomalies.
\end{itemize}	
 {The functions of mechanisms are described as follows:}
\begin{itemize}
	\item \textbf{\textit{Deep Anomaly Detection Mechanism:}} The deep anomaly detection mechanism is deployed in the edge devices, which can detect anomalies to reduce economic losses.
	\item \textbf{\textit{Gradient Compression Mechanism:}} The gradient compression mechanism is deployed in the edge devices, which can compress the local gradients to reduce the number of gradients exchanged between the edge devices  and  the cloud aggregator, thereby reducing communication overhead.
\end{itemize}
\subsection{Design Goals}
In this paper, our goal is to develop an on-device communication-efficient FL framework for deep anomaly detection in IIoT. First, the proposed framework needs to detect anomalies accurately in an unsupervised manner. The proposed framework uses an unsupervised AMCNN-LSTM model to detect anomalies. Second, the proposed framework can significantly improve communication efficiency by using a gradient compression mechanism. Third, the performance of the proposed framework is comparable to traditional FL frameworks.
\section{A Communication-Efficient On-device Deep Anomaly Detection Framework}
In this section, we first present the attention mechanism-based CNN-LSTM model. This model uses CNN to capture the fine-grained features of sensing time series data and uses LSTM module to accurately and timely detect anomalies. We then propose a deep gradient compression mechanism to further improve the communication efficiency of the proposed framework. %These methods are useful to implement in the following particular scenarios.
\begin{figure}[!t]
	\centering
	\includegraphics[width=1\linewidth]{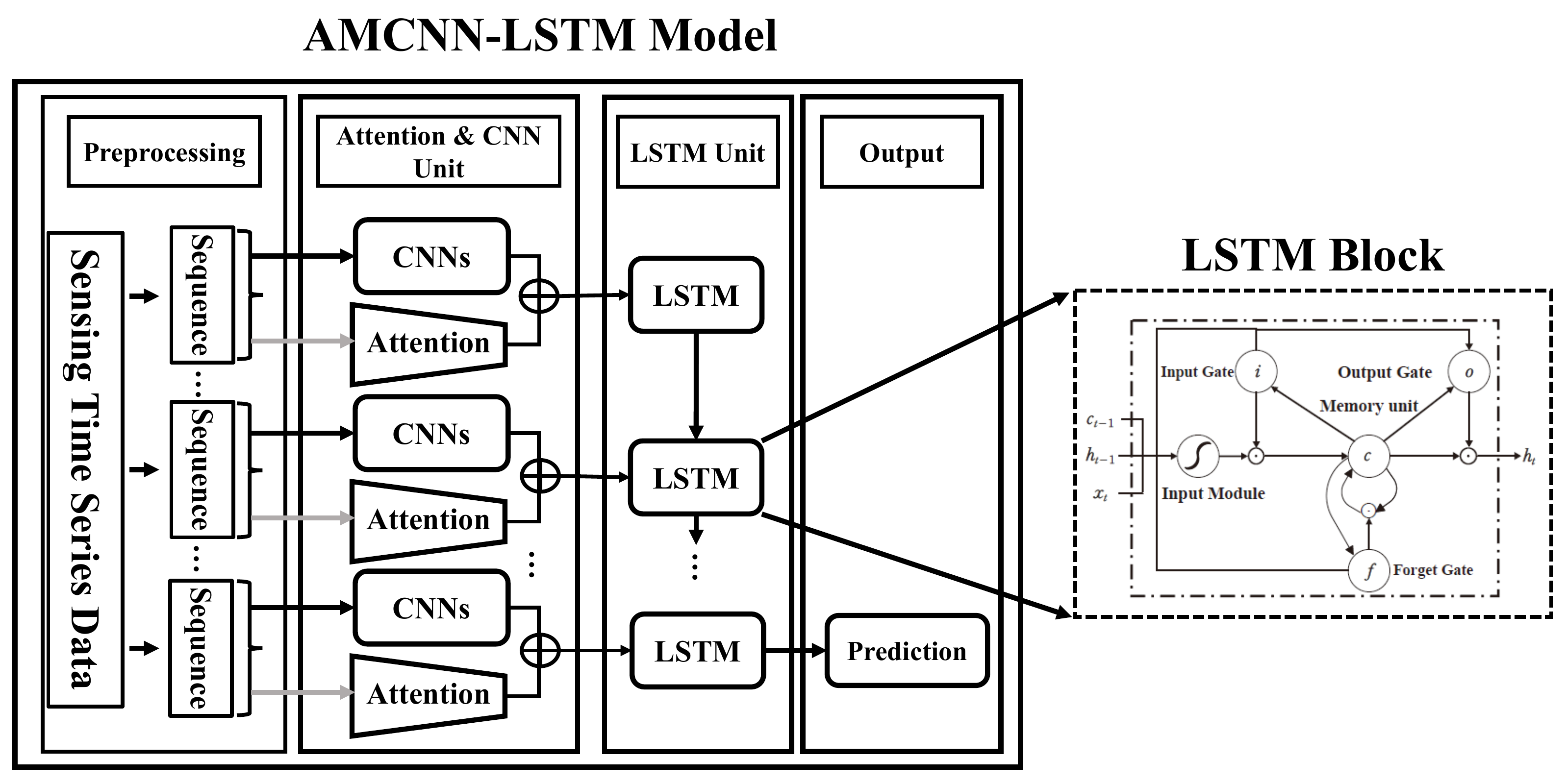}
	\caption{ {The overview of the attention mechanism-based CNN-LSTM Model.}}
	\label{fig-4}
	\vspace*{-5mm}
\end{figure}
\subsection{Attention Mechanism-based CNN-LSTM Model}
We present an unsupervised AMCNN-LSTM model including an input layer, an attention mechanism-based CNN unit, an LSTM unit, and an output layer shown in Fig. \ref{fig-4}. First, we use the preprocessed data as input to the input layer. Second, we use CNN to capture the fine-grained features of the input and utilize the attention mechanism to focus on the important features of CNN captured features. Third, we use the output of the attention mechanism-based CNN unit as the input of the LSTM unit and use LSTM to predict future time-series data. Finally, we propose an anomaly detection score to detect anomalies.

\textbf{Preprocessing:} We normalize the sensing time series data collected by the IIoT nodes into [0,1] to accelerate the model convergence.

\textbf{Attention Mechanism-based CNN Unit:} First, we introduce an attention mechanism in CNN unit to improve the focus on important features. In cognitive science, due to the bottleneck of information processing, humans will selectively focus on important parts of information while ignoring other visible information \cite{ref-35}. Inspired by the above facts, attention mechanisms are proposed for various tasks, such as computer vision and natural language processing \cite{ref-35,ref-36,ref-37}. Therefore, the attention mechanism can improve the performance of the model by paying attention to important features. The formal definition of the attention mechanism is given as follows:
\begin{equation}
\begin{array}{c}
{e_i} = a({\bf{u}},{{\bf{v}}_i})\; (\mathrm{Compute\;Attention\;Scores}),\\
{\alpha _i} = \frac{{{e_i}}}{{\sum\nolimits_i {{e_i}} }}\; (\mathrm{Normalize}),\\
c = \sum\limits_i {{\alpha _i}{{\bf{v}}_i}}\; (\mathrm{Encode}),
\end{array}
\end{equation}
where $\bf{u}$ is the matching feature vector based on the current task and is used to interact with the context, ${\bf{v}}_i$ is the feature vector of a timestamp in the time series, $e_i$ is the  unnormalized attention score, ${\alpha _i}$ is the normalized attention score, and $c$ is the context feature of the current timestamp calculated based on the attention score and feature sequence $\bf{v}$. In most instances, ${e_i} = {{\bf{u}}^T}W{\bf{v}}$, where $W$ is the weight matrix.

Second, we use CNN unit to extract fine-grained features of time series data. The CNN module is formed by stacking multiple layers of one-dimensional (1-D) CNN, and each layer includes a convolution layer, a batch normalization layer, and a non-linear layer. Such modules implement sampling aggregation by using pooling layers and create hierarchical structures that gradually extract more abstract features through the stacking of convolutional layers. This module outputs $m$ feature sequences of length $n$, and the size can be expressed as ($n \times m$). To further extract significant time-series data features, we propose a parallel feature extraction branch by combining the attention mechanisms and CNN. The attention mechanism module is composed of feature aggregation and scale restoration. The feature aggregation part uses the stacking of multiple convolutions and pooling layers to extract key features from the sequence and uses a convolution kernel of size $1 \times 1$ to mine the linear relationship. The scale restoration part restores the key features to ($n \times m$), which is consistent with the size of the output features of CNN module, and then uses the sigmoid function to constrain the values to [0,1]. 

Third, we multiply element-wise the output features of CNN module and the output of the important features by the corresponding attention mechanism module. We assume that the sequence ${X^i} = \{ x_1^i,x_2^i, \cdots ,x_n^i\} (0 \le i < I)$. The output of the sequence ${X^i}$ processed by CNN module is represented by $W_{\mathrm{CNN}}$, and the output of the corresponding attention module is represented as $W_{\mathrm{attention}}$. We multiply the two outputs element by element, as follows:
\begin{equation}
W(i,c) = {W_{\mathrm{CNN}}}(i,c) \odot {W_{\mathrm{attention}}}(i,c),
\end{equation}
where $\odot$ represents element-wise multiplication, $i$ is the corresponding position of the time series in the feature layer, and $c$ is the channel. We use the final feature layer $W (i, c)$ as the input of LSTM block. 

We introduce the attention mechanism to expand the receptive field of the input, which allows the model to obtain more comprehensive contextual information, thereby learning the important features of the current local sequence. Furthermore, we use the attention module to suppress the interference of unimportant features to the model, thereby solving the problem that the model cannot distinguish the importance of the time series data features.

\textbf{LSTM Unit:} In this paper, we use a variant of a recurrent neural network, called LSTM, to support accurately predict the sensing time series data to detect anomalies, as shown in Fig. \ref{fig-4}. LSTM uses a well-designed ``gate'' structure to remove or add information to the state of the cell. The ``gate'' structure is a method of selectively passing information. LSTM cells include forget gates $f_t$, input gates $i_t$, and output gates $o_t$. The calculations on the three gate structures are defined as follows:
\begin{equation}
\begin{array}{l}
{f_t} = {\sigma _l}({W_f} \cdot [{h_{t - 1}},{x_t}] + {b_f}),\\
{i_t} = {\sigma _l}({W_i} \cdot [{h_{t - 1}},{x_t}] + {b_i}),\\
{{\tilde C}_t} = \tanh ({W_C} \cdot [{h_{t - 1}},{x_t}] + {b_C}),\\
{C_t} = {f_t}*{C_{t - 1}} + {i_t}*{{\tilde C}_t},\\
{o_t} = {\sigma _l}({W_o} \cdot [{h_{t - 1}},{x_t}] + {b_o}),\\
{h_t} = {o_t}*\tanh ({C_t}),
\end{array}
\end{equation}
where ${W_f},W_i,W_C, W_o$, and ${b_f},{b_i},{b_C},{b_o}$ are the weight matrices and the bias vectors for input vector $x_t$ at time step $t$, respectively. $\sigma_l$ is the activation function, $*$ represents element-wise multiplication of a matrix, $C_t$ represents the cell state, $h_{t-1}$ is the state of the hidden layer at time step $t-1$, and $h_t$ is the state of the hidden layer at time step $t$.

\textbf{Anomaly Detection:} We use AMCNN-LSTM model to predict real-time and future sensing time series data in different edge devices:
\begin{equation}
[x_{n - T + 1}^i,x_{n - T + 2}^i, \cdots ,x_n^i]\mathop  \to \limits^{f( \cdot )} [x_{n + 1}^i,x_{n + 2}^i, \cdots ,x_{n + T}^i],
\end{equation}
where $f( \cdot )$ is the prediction function. In this paper, we use LSTM unit for time series prediction. We use anomaly scores for anomaly detection, which is defined as follows:
\begin{equation}
{A_n} = {({\beta _n} - \mu )^T}{\sigma ^{ - 1}}({\beta _n} - \mu ),
\end{equation}
where $A_n$ is the anomaly score, ${\beta _n} = |x_n^i - x_n^{'i}|$ is the reconstruction error vector, and the error vectors ${\beta _n}$ for the time series in the sequences $X^i$ are used to estimate the parameters $\mu$ and $\sigma$ of a Normal distribution $\mathcal{N} (\mu ;\sigma )$ using Maximum Likelihood Estimation. 

In an unsupervised setting, when ${A_n} \ge \varsigma $ ($\varsigma  = \max {F_\theta } = \frac{{(1 + {\theta ^2}) \times P \times R}}{{{\theta ^2}P + R}}$), where $P$ is precision, $R$ is recall, and $\theta$ is the parameter, a point in a sequence can be predicted to be ``anomalous'', otherwise ``normal''. 

\subsection{Gradient Compression Mechanism}\label{dgc}
%We know that if the gradients reach 99.9\% sparsity, only the 0.1\% gradients with the largest absolute value are useful for model aggregation \cite{ref-11}. 
 {If the gradients reach 99.9\% sparsity, only the 0.1\% gradients with the largest absolute value are useful for model aggregation \cite{ref-11}. Therefore, we only need to aggregate the gradient with a larger absolute value to update the model. This way reduces the byte size of the gradient matrix, which can reduce the number of gradients exchanged between the device and the cloud to improve communication efficiency, especially for distributed machine learning systems.} Inspired by the above facts, we propose a gradient compression mechanism to reduce the gradients exchanged between the cloud aggregator and the edge devices. We expect that this mechanism can further improve the communication efficiency of the proposed framework. 

%Since the edge devices only send some gradients with large absolute values to the cloud aggregator, the edge devices use the local gradient accumulation scheme to accumulate the remaining small gradients in the edge device to avoid information loss caused by gradient sparsification.

 {When we choose a gradient with a larger absolute value, we will meet the following situations: (1) All gradient values in the gradient matrix are not greater than the given threshold; (2) There are some gradient values in the gradient matrix that are very close to the given threshold. If we set these gradients that do not meet the threshold requirements to 0, it will cause information loss. Therefore, the device uses a local gradient accumulation scheme to prevent information loss. Specifically, the cloud returns smaller gradients to the device instead of filtering the gradients. The device keeps the smaller gradient in the buffer and accumulates all the smaller gradients until it reaches a given threshold.} Note that we use D-SGD for iterative updates, and the loss function to be optimized is defined as follows:
\begin{equation}
F(\omega ) = \frac{1}{{{D_k}}}\sum\limits_{x \in {D_k}} f (x,\omega ),
\end{equation}
\begin{equation}
{\omega _{t + 1}} = {\omega _t} - \eta \frac{1}{{Nb}}\sum\limits_{k = 1}^N {\sum\limits_{x \in {{\mathcal {B}}_{k,t}}} \nabla  } f\left( {x, {\omega _t}} \right),
\end{equation}
where $F(\omega)$ is the loss function, $f (x,\omega )$ is the loss function for the local device, $\omega$ are the weights of the model, $N$ is the total edge devices, $\eta$ is the learning rate, ${{\mathcal {B}}_{k,t}}$ represents the data sample for the $t$-th round of training, and each local dataset size of $b$.

When the gradients' sparsification reaches a high value (e.g., 99\%), it will affect the model convergence. By following  \cite{ref-11,ref-34}, we use momentum correction and local gradient clipping to mitigate this effect. Momentum correction can make the accumulated small local gradients converge toward the gradients with a larger absolute value, thereby accelerating the model's convergence speed. Local gradient clipping is used to alleviate the problem of gradient explosions \cite{ref-11}.  {Next, we prove that local gradient accumulation scheme will not affect the model convergence: We assume that ${g^{(i)}}$ is the $i$-th gradient, ${u^{(i)}}$ denotes the sum of the gradients using the aggregation algorithm in \cite{ref-28}, ${v^{(i)}}$ denotes the sum of the gradients using the local gradient accumulation scheme, and $m$ is the rate of gradient descent. If the $i$-th gradient does not exceed threshold until the $(t-1)$-th iteration and triggers the model update, we have:}
\begin{equation}\label{eq-3}
u_{t-1}^{(i)}=m^{t-2} g_{1}^{(i)}+\cdots+m g_{t-2}^{(i)}+g_{t-1}^{(i)},
\end{equation}
\begin{equation}\label{eq-4}
v_{t-1}^{(i)}=\left(1+\cdots+m^{t-2}\right) g_{1}^{(i)}+\cdots+(1+m) g_{t-2}^{(i)}+g_{t-1}^{(i)},
\end{equation}
then we can update $\omega_{t}^{(i)}=w_{1}^{(i)}-\eta \times v_{t-1}^{(i)}$ and set $v_{t-1}^{(i)}=0$. If the $i$-th gradient reaches the threshold at the $t$-th iteration, model update is triggered, thus we have:
\begin{equation}\label{eq-5}
u_{t}^{(i)}=m^{t-1} g_{1}^{(i)}+\cdots+m g_{t-1}^{(i)}+g_{t}^{(i)},
\end{equation}
\begin{equation}\label{eq-6}
v_t^{(i)} = {m^{t - 1}}g_1^{(i)} +  \cdots  + mg_{t - 1}^{(i)} + g_t^{(i)}.
\end{equation}
Then we can update $\omega _{t + 1}^{(i)} = \omega _t^{(i)} - \eta  \times v_t^{(i)} = {\omega _1}^{(i)} - \eta  \times \left[ {\left( {1 +  \cdots  + {m^{t - 1}}} \right)g_1^{(i)} +  \cdots  + (1 + m)g_{t - 1}^{(i)} + g_t^{(i)}} \right]=w_{1}^{(i)}-\eta \times v_{t-1}^{(i)}$, so the result of using the local gradient accumulation scheme is consistent with the usage effect of the optimization algorithm in \cite{ref-28}.

The specific implementation phases of the gradient compression mechanism are given as follows:
 {
\begin{enumerate}[label=\roman*)]
	\item \textbf{\textit{Phase 1, Local Training:}} Edge devices use the local dataset to train the local model. In particular, we use the gradient accumulation scheme to accumulate local small gradients.
	\item \textbf{\textit{Phase 2, Gradient Compression:}} Each edge device uses Algorithm 1 to compress the gradients and upload sparse gradients (i.e., \textbf{only gradients larger than a threshold are transmitted.}) to the cloud aggregator. Note that the edge devices send the remaining local gradient to the cloud aggregator when the local gradient accumulation is greater than a threshold.
	\item \textbf{\textit{Phase 3, Gradient Aggregation:}} The cloud aggregator obtains the global model by aggregating sparse gradients and sends this global model to the edge devices.
\end{enumerate}
}
%\begin{enumerate}[label=\roman*)]
%	\item \textbf{\textit{Phase 1, Local Training:}} Edge devices use the local dataset to train the local model. Each device uploads the local updated gradients to the cloud aggregator. In particular, we use the gradient accumulation scheme to accumulate local small gradients.
%	\item \textbf{\textit{Phase 2, Gradient Compression:}} Each edge device uses Algorithm \ref{al-2} to compress the gradients and upload sparse gradients (i.e., \textbf{only gradients larger than a threshold are transmitted.}) to the cloud aggregator. Note that the edge devices send the remaining local gradient to the cloud aggregator when the local gradient accumulation is greater than a threshold.
%	\item \textbf{\textit{Phase 3, Gradient Aggregation:}} The cloud aggregator obtains the global model by aggregating sparse gradients and sends this global model to the edge devices.
%\end{enumerate}
The gradient compression algorithm is thus presented in Algorithm \ref{al-2}. 
%\begin{algorithm}[t]\label{al-2}
%	\caption{Gradient compression mechanism.}%算法名字
%	\LinesNumbered %要求显示行号
%	\KwIn{Edge devices $\mathcal{N} = \{ {n_1},{n_2},\cdots,{n_i}\} $, $B$ is the local mini-batch size, $D_k$ is the local dataset, $\eta$ is the learning rate, and the optimization function $\mathrm{SGD}$.}%输入参数
%	\KwOut{$\omega$.}%输出
%	Initialize $ \omega_{t}$\;
%	${g^k} \leftarrow 0$\;
%	\For{$t = 0,1, \cdots $}
%	{
%		$g_t^k \leftarrow g_{t - 1}^k$\;
%		\For{$i = 1,2, \cdots $}
%		{
%			Sample data $x$ from $D_k$\;
%			$g_t^k \leftarrow g_t^k + \frac{1}{{NB}}\nabla f(x;{\omega _t})$\;
%		}
%	}
%	\If{Gradient Clipping}
%	{
%		$g_t^k \leftarrow \mathrm{Local\_Gradient\_Clipping}\,(g_t^k)$\;
%	}
%	\ForEach{$g_t^{{k_j}} \in \{ g_t^k\} $ and $j = 1,2, \cdots $}
%	{
%		$\mathrm{Thr} \leftarrow |\mathrm{Top}\,\rho \%\, of\,\{ g_t^k\} |$\;
%		\If{$|g_t^{{k_j}}| > \mathrm{Thr}$}
%		{
%			Send this gradient to the cloud aggregator\;
%			Send the remaining gradients to the edge device\;
%		}
%		\ElseIf{When accumulated local gradient $> \mathrm{Thr}$}
%		{
%			Send this gradient to the cloud aggregator\;
%		}
%		All-reduce $g_t^k:{g_t} \leftarrow \sum\nolimits_{k = 1}^N {(\mathrm{sparse}}\, \tilde g_t^k)$\;
%		$\omega_{t+1} \leftarrow \mathrm{SGD}\left(\omega_{t}, g_{t}\right)$.
%	}
%	\Return $\omega $.
%\end{algorithm}
\begin{algorithm}[t]\label{al-2}
 {	\caption{Gradient compression mechanism on edge node $k$.}%算法名字
	\LinesNumbered %要求显示行号
	\KwIn{$\mathcal{G}=  \{ {g^1},{g^2}, \ldots ,{g^k}\}$ is the edge node's gradient, $B$ is the local mini-batch size, $\mathcal{D}_k$ is the local dataset, $\eta$ is the learning rate, $f( \cdot , \cdot )$ is the edge node's loss function, and the optimization function $\mathrm{SGD}$.}%输入参数
	\KwOut{Parameter $\omega$.}%输出
	Initialize parameter $ \omega_{t}$\;
	${g^k} \leftarrow 0$\;
	\For{$t = 0,1, \cdots $}
	{
		$g_t^k \leftarrow g_{t - 1}^k$\;
		\For{$i = 1,2, \cdots $}
		{
			Sample data $x$ from $\mathcal{D}_k$\;
			$g_t^k \leftarrow g_{t-1}^k + \frac{1}{{|\mathcal{D}_k|B}}\nabla f(x;{\omega _t})$\;
		}
	}
	\If{Gradient Clipping}
	{
		$g_t^k \leftarrow \mathrm{Local\_Gradient\_Clipping}\,(g_t^k)$\;
	}
	\ForEach{$g_t^{{k_j}} \in \{ g_t^k\} $ and $j = 1,2, \cdots $}
	{
		$\mathrm{Thr} \leftarrow |\mathrm{Top}\,\rho \%\, of\,\{ g_t^k\} |$\;
		\If{$|g_t^{{k_j}}| \ge \mathrm{Thr}$}
		{
			Send this gradient to the cloud aggregator\;
		}
		\If{$|g_t^{{k_j}}| < \mathrm{Thr}$}
		{
			The edge node $k$ uses the local gradient accumulation scheme to accumulate gradients until the gradient reaches $\mathrm{Thr}$;
		}
		Aggregate $g_t^k:{g_t} \leftarrow \sum\nolimits_{k = 1}^N {(\mathrm{sparse}}\, \tilde g_t^k)$\;
		$\omega_{t+1} \leftarrow \mathrm{SGD}\left(\omega_{t}, g_{t}\right)$.
	}
	\Return $\omega $.}
\end{algorithm}

\section{Experiments}
%In this section, the proposed framework is applied to four real-world datasets, i.e., power demand, space shuttle, ECG, and engine (see \cite{ref-39} and Table \ref{tb-1}, where $X$, $X_n$, and $X_a$ is a number of original sequences, normal subsequences, and anomalous subsequences, respectively.) for performance demonstration. 
 {In this section, the proposed framework is applied to four real-world datasets, i.e., power demand \footnote{\href{https://archive.ics.uci.edu/ml/datasets/Individual+household+electric+power+consumption}{https://archive.ics.uci.edu/ml/datasets/}} , space shuttle \footnote{\href{https://archive.ics.uci.edu/ml/datasets/Statlog+(Shuttle)}{https://archive.ics.uci.edu/ml/datasets/Statlog+(Shuttle)}}, ECG \footnote{\href{https://physionet.org/about/database/}{https://physionet.org/about/database/}}, and engine \footnote{\href{https://archive.ics.uci.edu/ml/datasets.php}{https://archive.ics.uci.edu/ml/datasets.php}} for performance demonstration. These datasets are time series datasets collected by different types of sensors from different fields \cite{ref-2}. For example, the power demand dataset is composed of electricity consumption data recorded by the electricity meter. There are normal subsequences and anomalous subsequences in these datasets. As shown in Table \ref{tb-1}, $X$, $X_n$, and $X_a$ is a number of original sequences, normal subsequences, and anomalous subsequences, respectively. For the power demand dataset, the anomalous subsequences indicate that the electricity meter has failed or stop working. Therefore, we need to use these datasets to train a FL model that can detect anomalies.}
We divide all datasets into a training set and a test set in a 7: 3 ratio. We implement the proposed framework by using Pytorch and PySyft \cite{ref-38}. The experiment is conducted on a virtual workstation with the Ubuntu 18.04 operation system, Intel (R) Core (TM) i5-4210M CPU, 16GB RAM, 512GB SSD. 
\begin{table}[!t],
	\centering
	\caption{Details of four real-world datasets}
	\begin{tabular}{ccccc} % 控制表格的格式
		\toprule
		Datasets  &Dimensions	&$X$	&$X_n$	&$X_a$		\\
		\midrule
		Power Demand &1 &1  &45  &6   \\
		Space Shuttle      &1	&3	&20   &8	\\
		ECG      &1	&1	&215   &1	\\
		Engine       &12	&30	&240   &152		\\
		\bottomrule
	\end{tabular}
	\label{tb-1}
	\vspace*{-5mm}
\end{table}
\subsection{Evaluation Setup}
In this experiment, to determine the hyperparameter $\rho$ of the gradient compression mechanism, we first apply a simple CNN network (i.e., CNN with 2 convolutional layers followed by 1 fully connected layer) in the proposed framework to perform the classification task on MNIST and CIFAR-10 dataset. The pixels in all datasets are normalized into [0,1]. During the simulation, the number of edge devices is  $N = 10$, the learning rate is  $\eta = 0.001$, the training epoch is  $E= 1000$, the mini-batch size is  $B = 128$, and we follow reference \cite{ref-39} and set $\theta$  as 0.05.

We adopt Root Mean Square Error (RMSE) to indicate the performance of AMCNN-LSTM model as follows:
\begin{equation}
\mathrm{RMSE} = {[\frac{1}{n}\sum\limits_{i = 1}^n {(|{y_i} - {{\hat y}_p}|} )^2}{]^{\frac{1}{2}}},
\end{equation}
where $y_i$ is the observed sensing time series data, and ${\hat y}_p$ is the predicted sensing time series data.
\begin{figure}[!t]
	\centering
	\includegraphics[width=0.8\linewidth]{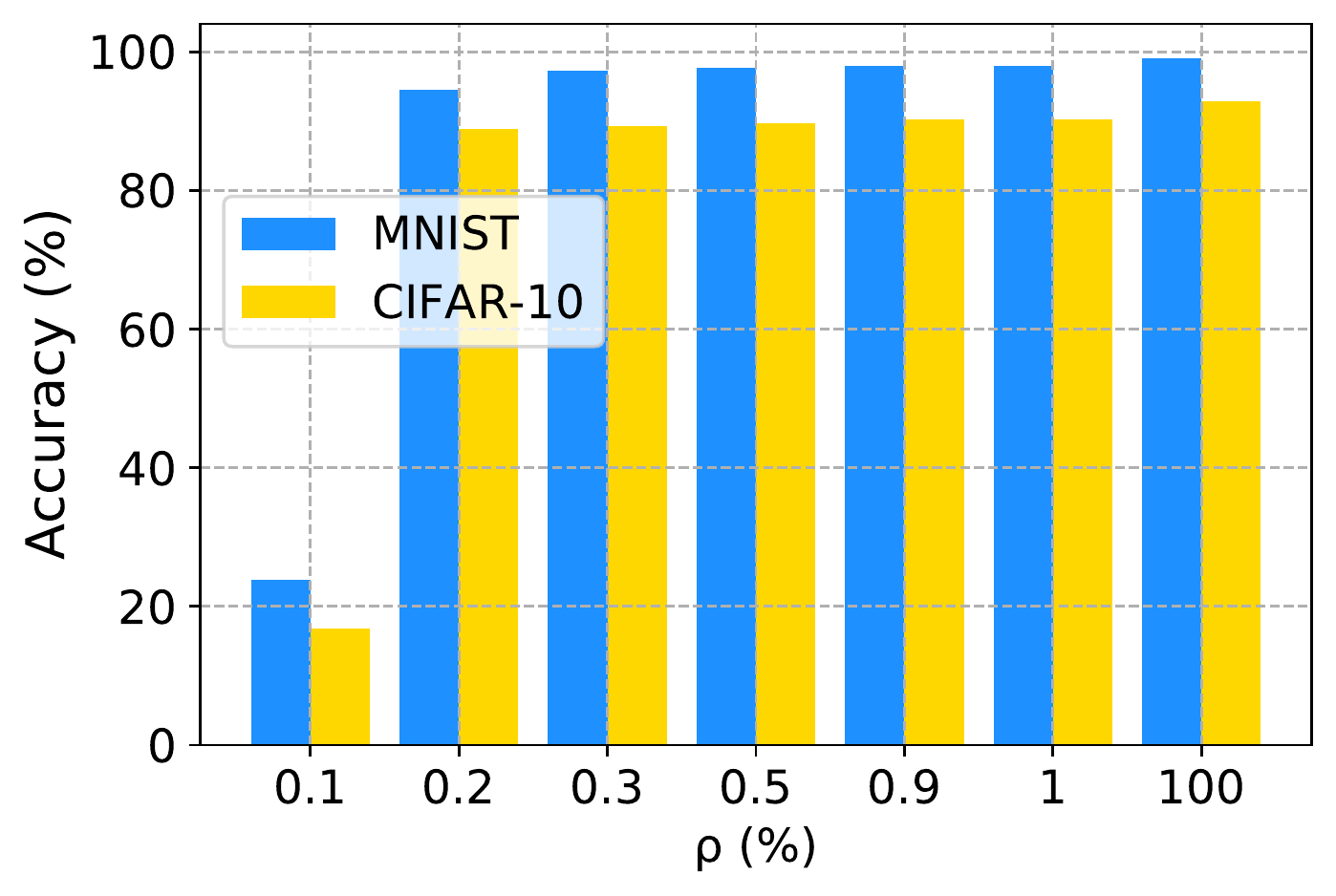}
	\caption{ {The accuracy of the proposed framework with different $\rho$ on MNIST and CIFAR-10 datasets.}}
	\label{fig-5}
\end{figure}
\subsection{ {Hyperparameters Selection of the Proposed Framework}}\label{gcm}
In the context of deep gradient compression scheme, proper hyperparameter selection, i.e., a threshold of absolute gradient value, is a notable factor that determines the proposed framework performance. In this section, we investigate the performance of the proposed framework with different thresholds and try to find a best-performing threshold for it. In particular, we employ $\rho \in \{0.1,0.2,0.3,0.5,0.9,1,100\}$ to adjust the best threshold of the proposed framework.  We use MNIST and CIFAR-10 datasets to evaluate the performance of the proposed framework with the selected threshold. As shown in Fig. \ref{fig-5}, we observe that the larger $\rho$, the better the performance of the proposed framework. For MNIST task, the results show that when $\rho = 0.3$, the accuracy is 97.25\%; when $\rho = 100$, the accuracy is 99.08\%. This means that the model increases gradient size by about 300 times, but the accuracy is only improved by 1.83\%. Furthermore, we observe a trade-off between the gradient threshold and accuracy. Therefore, to achieve a good trade-off between the gradient threshold and accuracy, we choose $\rho = 0.3$ as the best threshold of our scheme. 
\begin{figure}[!t]
	\centering
	\includegraphics[width=0.8\linewidth]{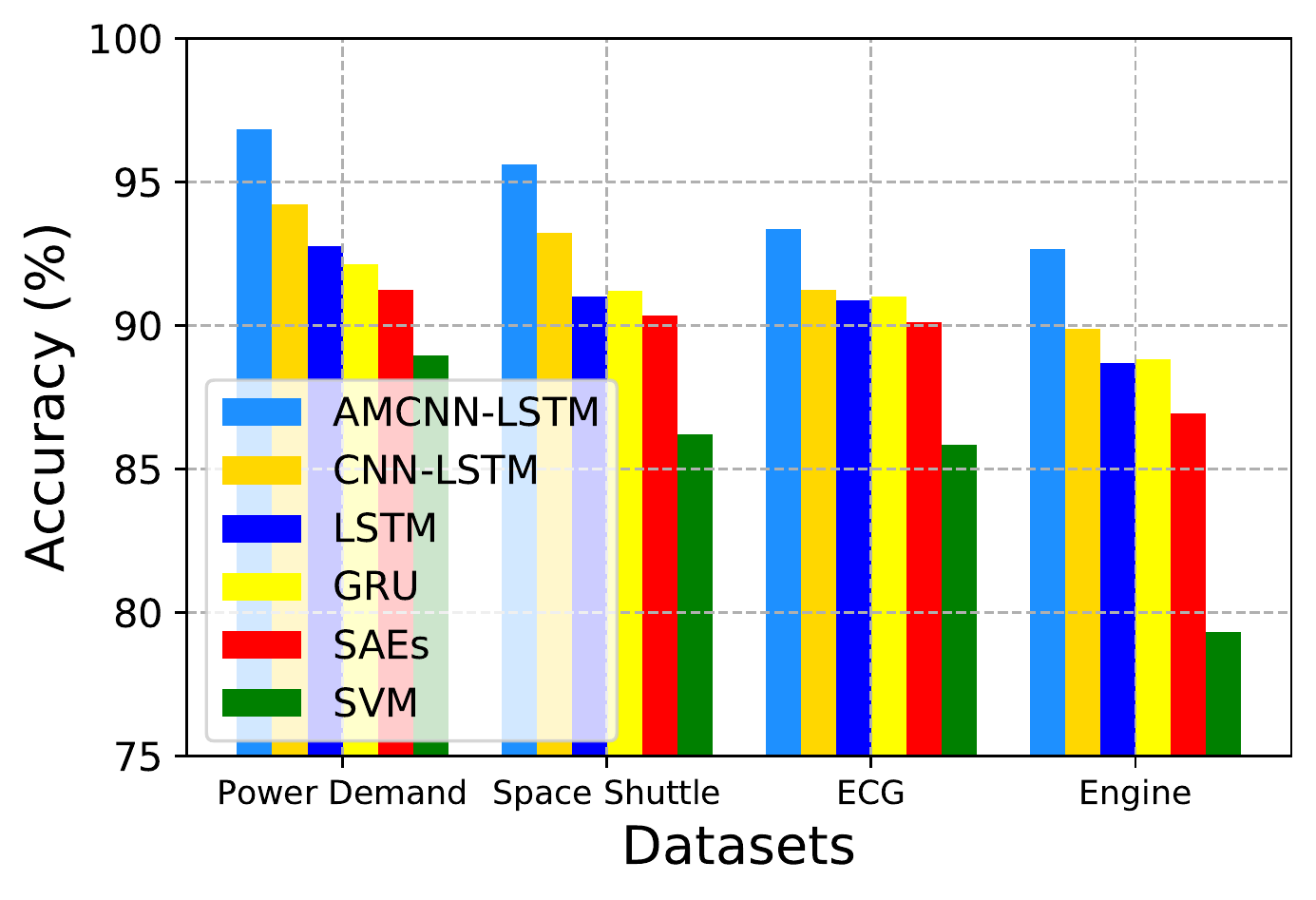}
	\caption{Performance comparsion of detection accuracy for AMCNN-LSTM, CNN-LSTM, LSTM, GRU, SAEs, and SVM on different datasets: power demand, space shuttle, ECG, and engine.}
	\label{fig-6}
	\vspace*{-5mm}
\end{figure}
\begin{figure}[!t]
	\centering
	\includegraphics[width=0.8\linewidth]{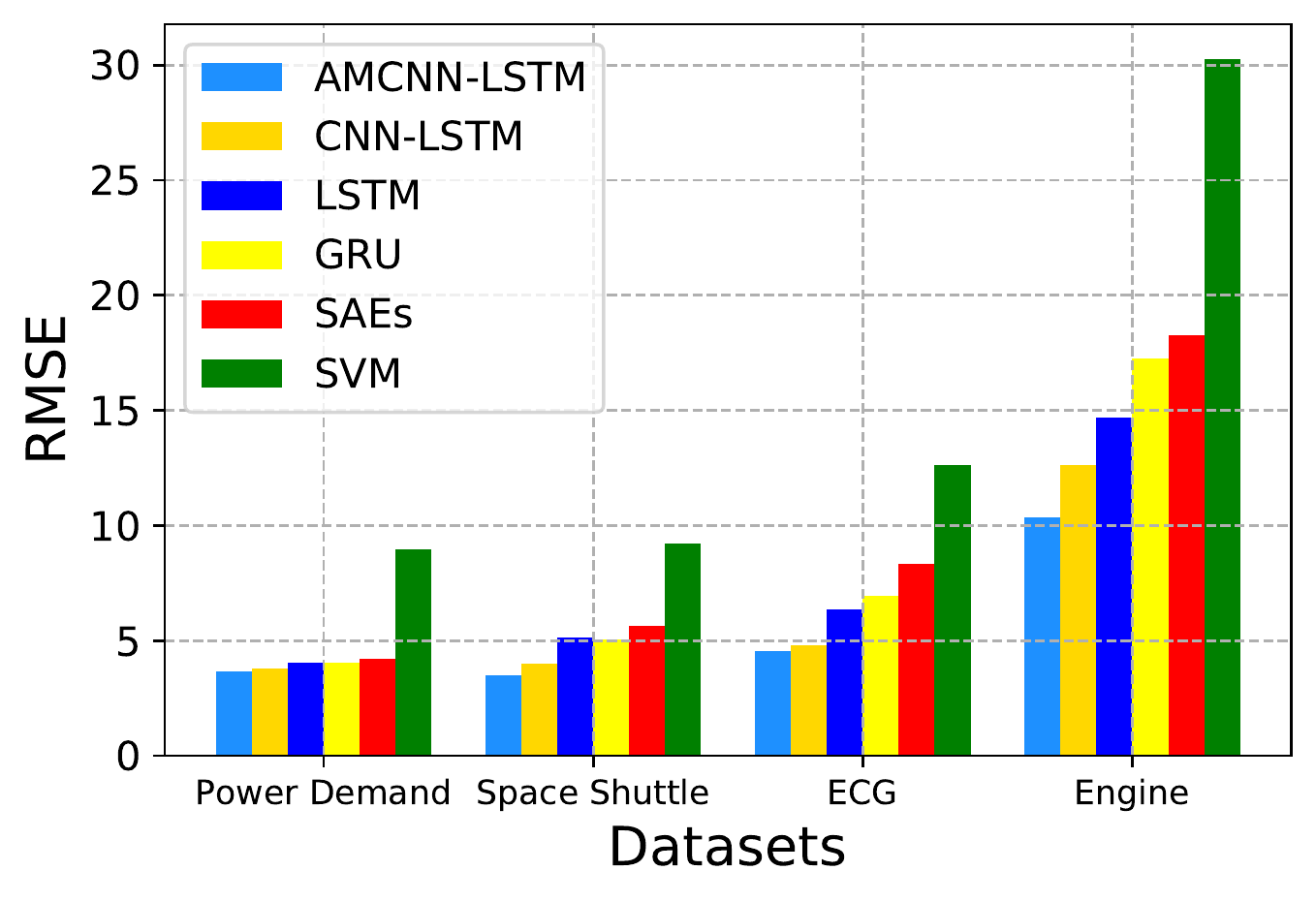}
	\caption{Performance comparsion of RMSE for AMCNN-LSTM, CNN-LSTM, LSTM, GRU, SAEs, and SVM on different datasets: power demand, space shuttle, ECG, and engine.}
	\label{fig-7}
	\vspace*{-5mm}
\end{figure}
\subsection{Performance of the Proposed Framework}
We compared the performance of the proposed model with that of CNN-LSTM \cite{ref-40}, LSTM \cite{ref-39}, Gate Recurrent Unit (GRU) \cite{ref-41}, Stacked Auto Encoders (SAEs) \cite{ref-42}, and Support Machine Vector (SVM) \cite{ref-43} method with an identical simulation configuration. Among these competing methods, AMCNN-LSTM is a FL-based model, and the rest of the methods are centralized ones. All models are popular DAD models for general anomaly detection applications. We evaluate these models on four real-world datasets, i.e., power demand, space shuttle, ECG, and engine. 

First, we compare the accuracy of the proposed model with competing methods in anomaly detection. We determine the $\max {F_\theta }$ and hyperparameter $\varsigma $ based on the accuracy and recall of the model on the training set. The hyperparameters $\varsigma $ of the dataset power demand, space shuttle, ECG, and engine are 0.75, 0.80, 0.80, and 0.60. In Fig. \ref{fig-6}, experimental results show that the proposed model achieves the highest accuracy on all four datasets. For example, for the dataset power demand, the accuracy of AMCNN-LSTM model is 96.85\%, which is 7.87\% higher than that of SVM model. From the experimental results, AMCNN-LSTM has better robustness to different datasets. The reason is that we use the on-device FL framework to train and update the model, which can learn the time-series features from different edge devices as much as possible, thereby improving the robustness of the model. Furthermore, the FL framework provides opportunities for edge devices to update models in a timely manner. This helps the edge device owner to update the model on the edge devices in time.

Second, we need to evaluate the prediction error of the proposed model and the competing methods. As shown in Fig. \ref{fig-6}, experimental results show that the proposed model achieves the best performance on four real-world datasets. For the ECG dataset, RMSE of AMCNN-LSTM model is 63.9\% lower than that of  SVM model. The reason is that AMCNN-LSTM model uses AMCNN units to capture important fine-grained features and prevent memory loss and gradient dispersion problems. Memory loss and gradient dispersion problems often occur in encoder-decoder models such as LSTM and GRU models. Furthermore, the proposed model retains the advantages of LSTM unit in predicting time series data. Therefore, the proposed model can accurately predict time series data.

Therefore, the proposed model  not only accurately detects abnormalities, but also accurately predicts time series data.
\begin{figure}[!t]
	\centering
	\includegraphics[width=0.8\linewidth]{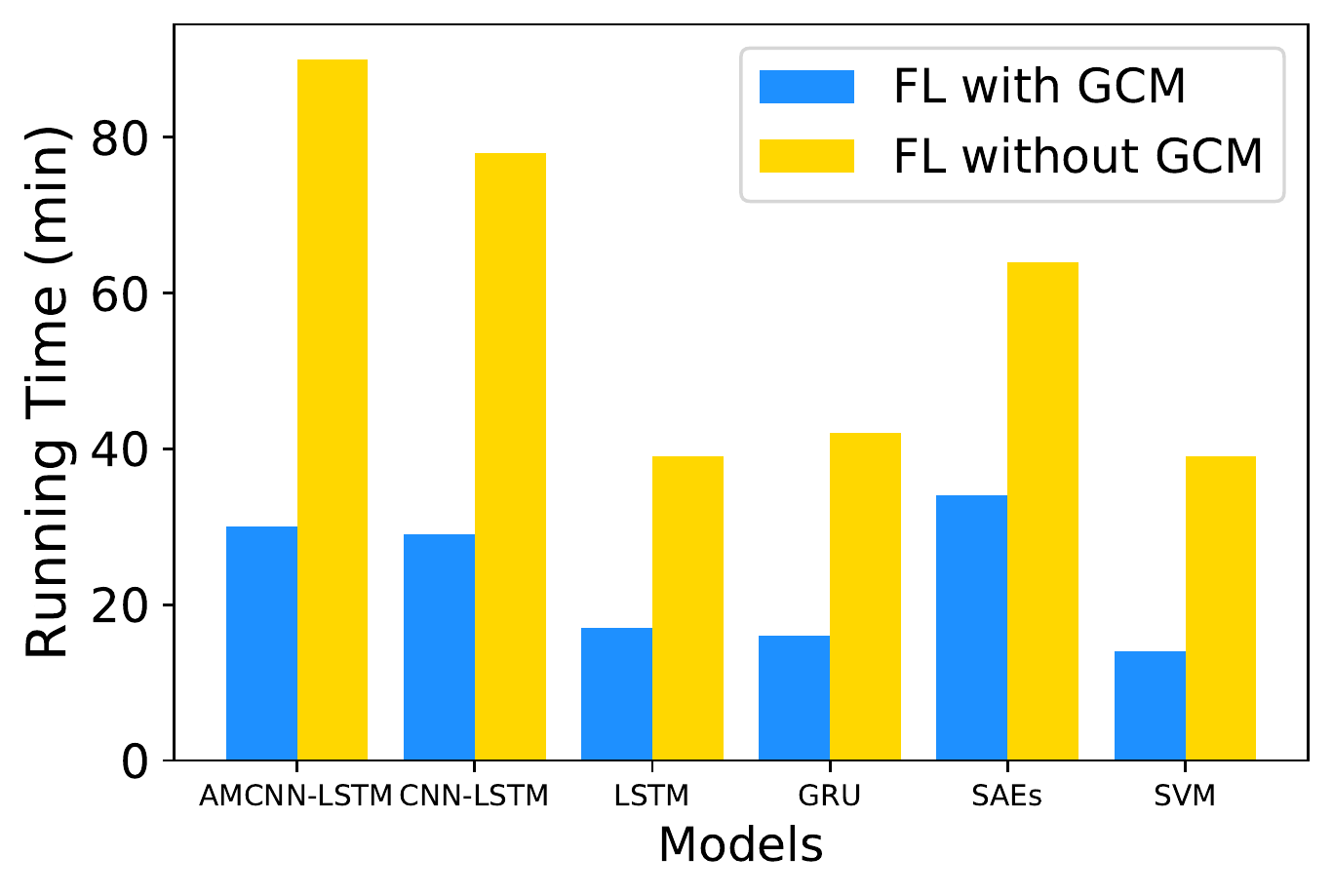}
	\caption{Comparison of communication efficiency between FL with GCM and FL without GCM with different models.}
	\label{fig-8}
	\vspace*{-5mm}
\end{figure}
\subsection{Communication Efficiency of the Proposed Framework}
In this section, we compare the communication efficiency between FL framework with the gradient compression mechanism (GCM) and the traditional FL framework without GCM. We apply the same model (i.e., AMCNN-LSTM, CNN-LSTM, LSTM, GRU, SAEs, and SVM) in the proposed framework and the traditional FL framework. Note that we fix the communication overhead of each round, so we can compare the running time of the model to compare the communication efficiency. In Fig. \ref{fig-8}, we show the running time of FL with GCM and FL without GCM using different models. As shown in Fig. \ref{fig-8}, we observe that the running time of FL framework with GCM is about 50\% that of the framework without GCM. The reason is that GCM can reduce the number of gradients exchanged between the edge devices and the cloud aggregator. In section \ref{dgc}, we show that GCM can compress the gradient by 300 times without compromising accuracy. Therefore, the proposed communication efficient framework is practical and effective in real-world applications.
\subsection{ {Discussion}}
 {Due to the trade-off between privacy and model performance, we will discuss the privacy analysis of the proposed framework in terms of data access and model performance:}
 {
\begin{itemize}
	\item \textbf{Data Access:} FL framework allows edge devices to keep the dataset locally and collaboratively learn deep learning models, which means that any third party cannot access the user's raw data. Therefore, the FL-based model can achieve anomaly detection without compromising privacy.
	\item \textbf{Model Performance:} Although the FL-based model can protect privacy, the model performance is still an important metric to measure the quality of the model. It can be seen from the experimental results that the performance of the proposed model is comparable to many advanced centralized machine learning models, such as CNN-LSTM, LSTM, GRU, and SVM model. In other words, the proposed model makes a good compromise between privacy and model performance.
\end{itemize}
}
\section{Conclusion}
In this paper, we propose a novel communication-efficient on-device FL-based deep anomaly detection framework for sensing time series data in IIoT. First, we introduce a FL framework to enable decentralized edge devices to collaboratively train an anomaly detection model, which can solve the problem of data islands. Second, we propose an attention mechanism-based CNN-LSTM (AMCNN-LSTM) model to accurately detect anomalies. AMCNN-LSTM model uses attention mechanism-based CNN units to capture important fine-grained features and prevent memory loss and gradient dispersion problems. Furthermore, this model retains the advantages of LSTM unit in predicting time series data. We evaluate the performance of the proposed model on four real-world datasets and compare it with CNN-LSTM, LSTM, GRU, SAEs, and SVM methods. The experimental results show that the AMCNN-LSTM model can achieve the highest accuracy on all four datasets. Third, we propose a gradient compression mechanism based on Top-\textit{k} selection to improve communication efficiency. Experimental results validate that this mechanism can compress the gradient by 300 times without losing accuracy. To the best of our knowledge, this is one of the pioneering work for deep anomaly detection by using  on-device  FL.

 {In the future, we will focus on researching privacy-enhanced FL frameworks and more robust anomaly detection models. The reason is that the FL framework is vulnerable to malicious attacks by malicious participants and a more robust model can be applied to a wider range of application scenarios.}

\ifCLASSOPTIONcaptionsoff
  \newpage
\fi

\begin{IEEEbiography}[{\includegraphics[width=1in,height=1.25in,clip,keepaspectratio]{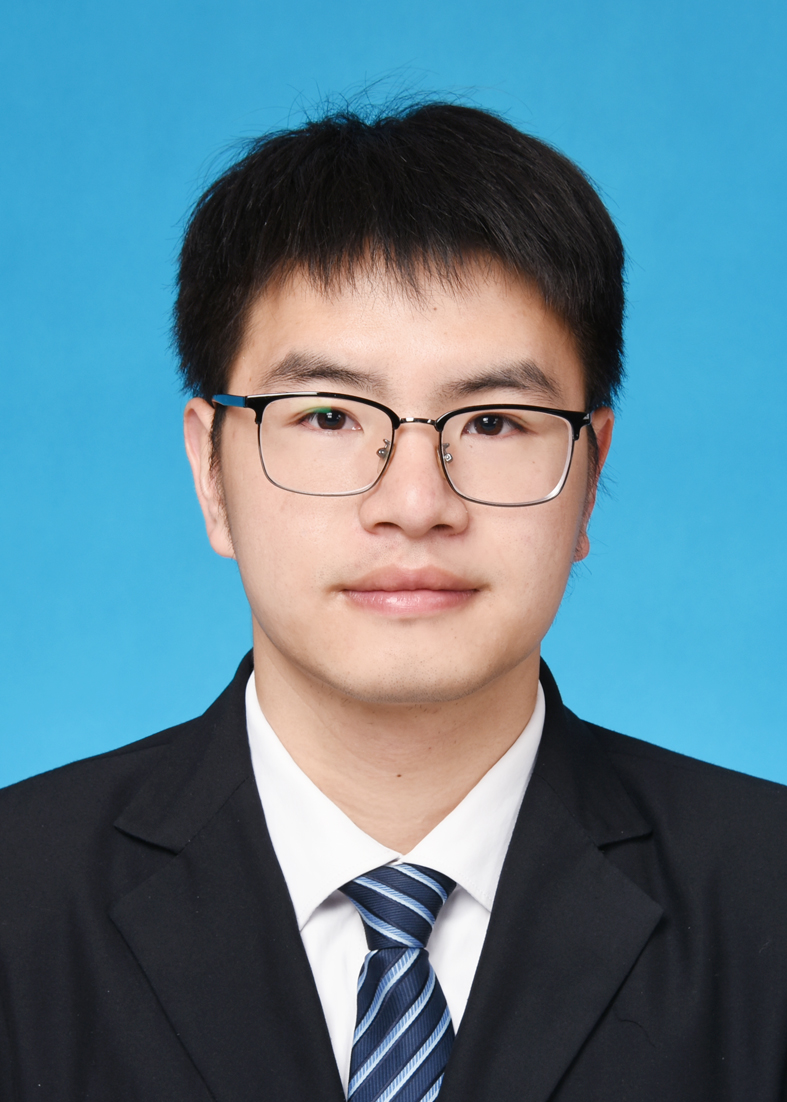}}]
{Yi Liu}(S'19) received the B.Eng. degree in network engineering from Heilongjiang University, Harbin, China, in 2019. He is currently pursuing a Ph.D. degree at the Faculty of Information Technology, Monash University, Melbourne, Australia. His research interests include security \& privacy, federated learning, edge computing, and blockchain.
\end{IEEEbiography}

\begin{IEEEbiography}[{\includegraphics[width=1in,height=1.25in,clip,keepaspectratio]{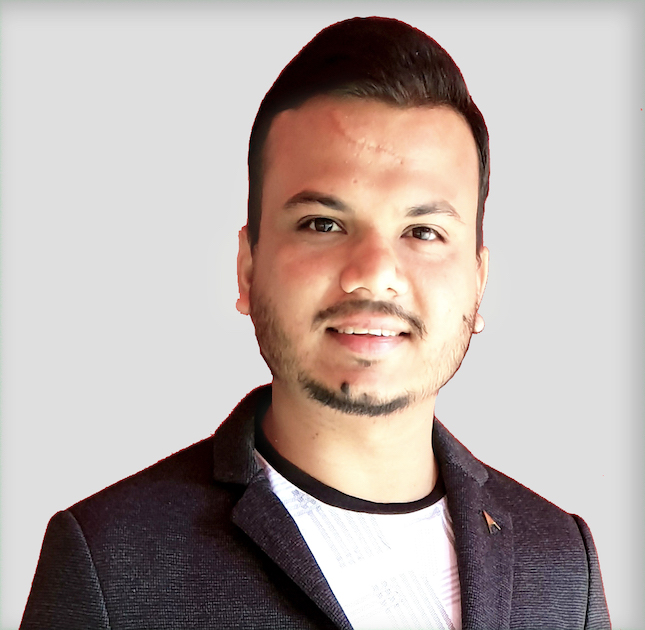}}]
{Sahil Garg}  (S'15, M'18) is a postdoctoral research fellow at École de technologie supérieure, Université du Québec, Montréal, Canada. He received his Ph.D. degree from the Thapar Institute of Engineering and Technology, Patiala, India, in 2018. He has many research contributions in the area of machine learning, big data analytics, security and privacy, the Internet of Things, and cloud computing. He has over 50 publications in high ranked journals and conferences, including 25+ IEEE transactions/journal papers. He received the IEEE ICC best paper award in 2018 in Kansas City, Missouri. He serves as the Managing Editor of Springer’s Human-Centric Computing and Information Sciences journal. He is also an Associate Editor of IEEE Network, IEEE System Journal, Elsevier’s Applied Soft Computing, Future Generation Computer Systems, and Wiley’s International Journal of Communication Systems. In addition, he also serves as a Workshops and Symposia Officer of the IEEE ComSoc Emerging Technology Initiative on Aerial Communications. He has guest edited a number of Special Issues in top-cited journals including IEEE T-ITS, IEEE TII, the IEEE IoT Journal, IEEE Network, and Future Generation Computer Systems. He serves/served as the Workshop Chair/Publicity Co-Chair for several IEEE/ACM conferences including IEEE INFOCOM, IEEE GLOBECOM, IEEE ICC, ACM MobiCom, and more. He is a member of ACM.
\end{IEEEbiography}

\begin{IEEEbiography}[{\includegraphics[width=1in,height=1.25in,clip,keepaspectratio]{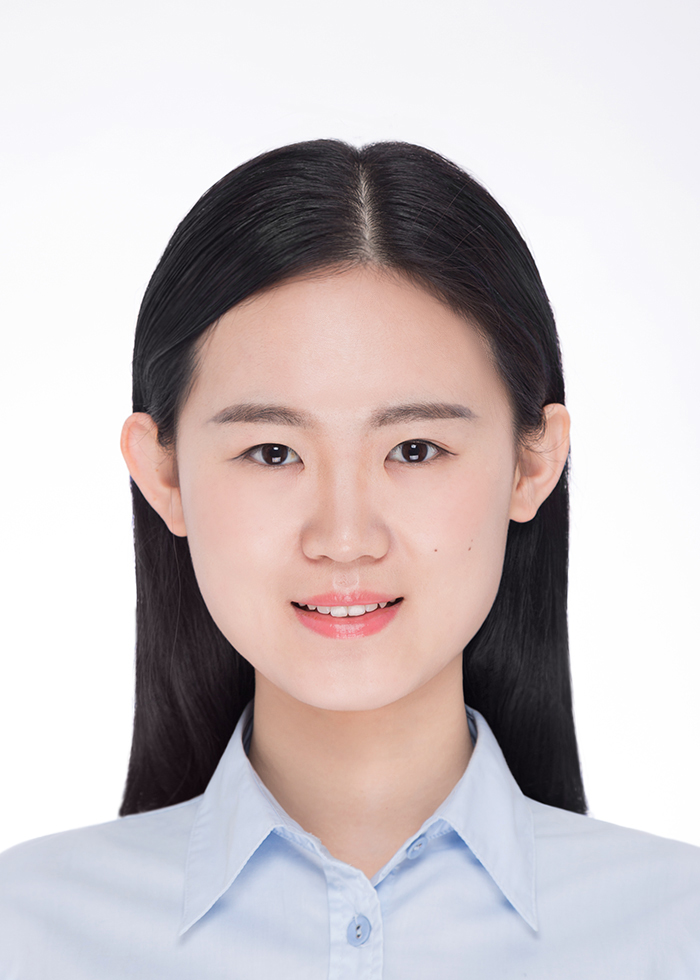}}]
{Jiangtian Nie} received her B.Eng degree with honors in Electronics and Information Engineering from Huazhong University of Science and Technology, Wuhan, China, in 2016. She is currently working towards the Ph.D. degree with ERI@N in the Interdisciplinary Graduate School, Nanyang Technological University, Singapore. Her research interests include incentive mechanism design in crowdsensing and game theory.
\end{IEEEbiography}

\begin{IEEEbiography}[{\includegraphics[width=1in,height=1.25in,clip,keepaspectratio]{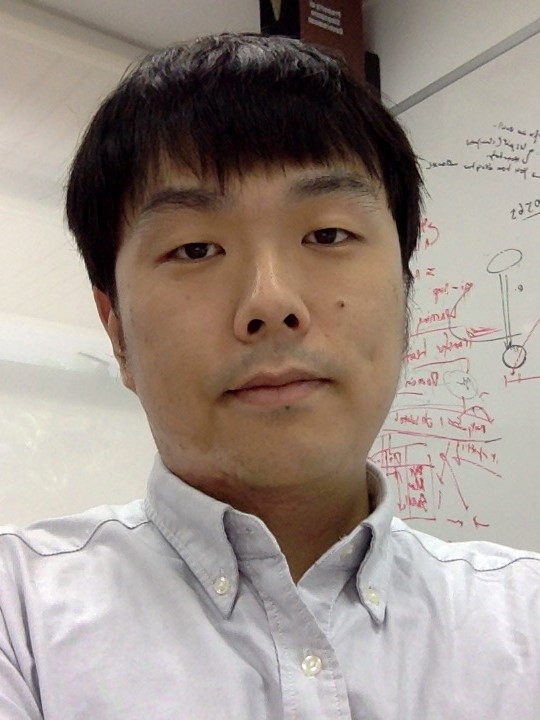}}]
{Yang Zhang} (M’11) is currently an associate professor at the School of Computer Science and Technology, Wuhan University of Technology, China. He received B. Eng. and M. Eng. from Beijing University of Aeronautics and Astronautics in 2008 and 2011, respectively, and obtained a Ph.D. degree in Computer Engineering from Nanyang Technological University (NTU), Singapore, in 2015. He is an associate editor of EURASIP Journal on Wireless Communications and Networking, and a technical committee member of Computer Communications (Elsevier). He is also an advisory expert member of Shenzhen FinTech Laboratory, China. His current research interests include: market-oriented modeling for network resource allocation, multiple agent machine learning, and deep reinforcement learning in network systems.
\end{IEEEbiography}

\begin{IEEEbiography}[{\includegraphics[width=1in,height=1.25in,clip,keepaspectratio]{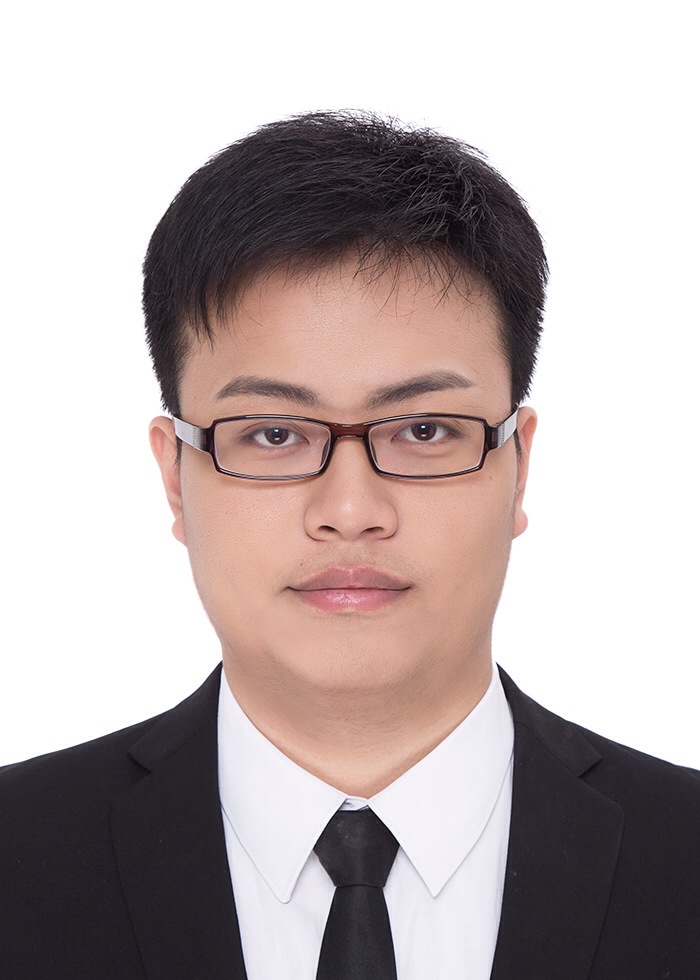}}]
{Zehui Xiong}(S'17, M'20) is currently a researcher with Alibaba-NTU Singapore Joint Research Institute, Singapore. He obtained the B.Eng degree with the highest honors in Telecommunications Engineering at Huazhong University of Science and Technology, Wuhan, China, in Jul 2016. He received the Ph.D. degree in Computer Science and Engineering at Nanyang Technological University, Singapore, in Apr 2020. He was a visiting scholar at Princeton University and University of Waterloo. His research interests include resource allocation in wireless communications, network games and economics, blockchain, and edge intelligence. He has won several Best Paper Awards including IEEE WCNC 2020, and IEEE Vehicular Technology Society Singapore Best Paper Award in 2019. He is an Editor for Elsevier Computer Networks and Elsevier Physical Communication, and an Associate Editor for IET Communications. He serves as a Guest Editor for IEEE Transactions on Cognitive Communications and Networking, IEEE Open Journal of Vehicular Technology, and EURASIP Journal on Wireless Communications and Networking. He is the recipient of the Chinese Government Award for Outstanding Students Abroad in 2019, and NTU SCSE Outstanding PhD Thesis Runner-Up Award in 2020.
\end{IEEEbiography}

\begin{IEEEbiography}
	[{\includegraphics[width=1in,height=1.25in,clip,keepaspectratio]{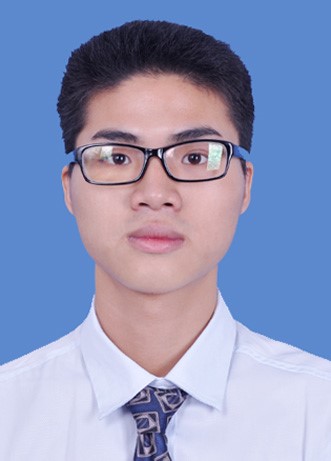}}]
	{Jiawen Kang} received the M.S. degree from the Guangdong University of Technology, China, in 2015, and the Ph.D. degree at the same school in 2018. He is currently  a postdoc at Nanyang Technological University, Singapore. His research interests mainly focus on blockchain, security and privacy protection in wireless communications and networking.
\end{IEEEbiography}

~\\
\footnotesize{ \textbf{\textbf{M. Shamim Hossain}} (SM'19) is a Professor at the Department of Software Engineering, College of Computer and Information Sciences, King Saud University, Riyadh, Saudi Arabia. He is also an adjunct professor at the School of Electrical Engineering and Computer Science, University of Ottawa, Canada. He received his Ph.D. in Electrical and Computer Engineering from the University of Ottawa, Canada. His research interests include cloud networking, smart environment (smart city, smart health), AI, deep learning, edge computing, Internet of Things (IoT), multimedia for health care, and multimedia big data. He has authored and coauthored more than 250 publications including refereed journals, conference papers, books, and book chapters. Recently, he co-edited a book on ``Connected Health in Smart Cities", published by Springer.  He has served as cochair, general chair, workshop chair, publication chair, and TPC for over 12 IEEE and ACM conferences and workshops. Currently, he is the cochair of the 3rd IEEE ICME workshop on Multimedia Services and Tools for smart-health (MUST-SH 2020).  He is a recipient of a number of awards, including the Best Conference Paper Award and the 2016 ACM Transactions on Multimedia Computing, Communications and Applications (TOMM) Nicolas D. Georganas Best Paper Award. He is on the editorial board of the IEEE Transactions on Multimedia, IEEE Multimedia, IEEE Network, IEEE Wireless Communications, IEEE Access, Journal of Network and Computer Applications (Elsevier), and International Journal of Multimedia Tools and Applications (Springer). He also presently serves as a lead guest editor of and Multimedia systems Journal. He serves/served as a guest editor of IEEE Communications Magazine, IEEE Network, ACM Transactions on Internet Technology, ACM Transactions on Multimedia Computing, Communications, and Applications (TOMM), IEEE Transactions on Information Technology in Biomedicine (currently JBHI), IEEE Transactions on Cloud Computing, Multimedia Systems, International Journal of Multimedia Tools and Applications (Springer), Cluster Computing (Springer), Future Generation Computer Systems (Elsevier). He is a senior member of both the IEEE, and ACM.}

\end{document}